\documentclass[runningheads]{llncs}
\usepackage{amsmath}
\usepackage{amssymb}
\usepackage{graphicx}
\usepackage{subcaption}
\usepackage{cancel}
\usepackage{array}
\usepackage{makecell}
\usepackage{multirow}
\usepackage{hyperref}
\usepackage{xcolor}
\usepackage{tabularx}
\usepackage{float}

\newcommand{\relu}{\mbox{ReLU}}

\begin{document}
\title{Pruning artificial neural networks:\\ a way to find well-generalizing,\\high-entropy sharp minima}
\titlerunning{Pruning ANNs: a way to find well-generalizing, high-entropy sharp minima}
%
\author{Enzo Tartaglione\orcidID{0000-0003-4274-8298} \and
Andrea Bragagnolo \and \\
Marco Grangetto\orcidID{0000-0002-2709-7864}}
\authorrunning{E. Tartaglione et al.}
%
\institute{University of Torino, Computer Science dept., Torino TO 10149, Italy\\
\email{\{enzo.tartaglione, andrea.bragagnolo\}@unito.it}
}
\maketitle              
\begin{abstract}
Recently, a race towards the simplification of deep networks has begun, showing that it is effectively possible to reduce the size of these models with minimal or no performance loss. However, there is a general lack in understanding why these pruning strategies are effective.\\
In this work, we are going to compare and analyze pruned solutions with two different pruning approaches, one-shot and gradual, showing the higher effectiveness of the latter. In particular, we find that gradual pruning allows access to narrow, well-generalizing minima, which are typically ignored when using one-shot approaches. In this work we also propose PSP-entropy, a measure to understand how a given neuron correlates to some specific learned classes. Interestingly, we observe that the features extracted by iteratively-pruned models are less correlated to specific classes, potentially making these models a better fit in transfer learning approaches.  

\keywords{Pruning \and Sharp minima \and Entropy \and Post synaptic potential \and Deep learning.}
\end{abstract}

\section{Introduction}
\label{sec:intro}
Artificial neural networks (ANNs) are nowadays one of the most studied algorithm used to solve a huge variety of tasks. Their success comes from their ability to learn from examples, not requiring any specific expertise and using very general learning strategies. The use of GPUs (and, recently, TPUs) for training ANNs gave a decisive kick to their large-scale deploy.\\
However, many deep models share a common obstacle: the large number of parameters, which allows their successful training~\cite{ba2014deep,denton2014exploiting}, determines in turn a large number of operations at inference time, preventing efficient deployment to mobile and cheap embedded devices.\\
In order to address this problem, a number of approaches have been proposed, like defining new, more efficient models~\cite{howard2017mobilenets}. Recently, a race to shrink the size of these ANN models has begun: the so-called \emph{pruning strategies} are indeed able to remove (or \emph{prune}) non-relevant parameters from pre-trained models, reducing the size of the ANN model, yet keeping a high generalization capability. On this topic, a very large amount of strategies have been proposed~\cite{frankle2018lottery,liu2018rethinking,lobster,tartaglione2018sensitivity} from which we can identify two main classes:
\begin{itemize}
    \item one-shot strategies: which are able to prune parameters using very fast, greedy approaches;
    \item gradual strategies: much slower than one-shot approaches, potentially they can achieve higher \emph{compression~rates} (or in other words, they promise to prune more parameters at the cost of higher computational complexity).
\end{itemize}
In such a rush, however, an effort into a deeper understanding on potential properties of such sparse architectures has been mostly set aside: is there a specific reason for which we are able to prune many parameters with minimal or no generalization loss? Are one-shot strategies enough to match gradual pruning approaches? Is there any hidden property behind these sparse architectures?\\
In this work, we first compare one-shot pruning strategies to their gradual counterparts, investigating the eventual benefits of having a much more computationally-intensive sparsifying strategy. Then, we shine a light on some local properties of minima achieved using the two different pruning strategies and finally, we propose \emph{PSP-entropy}, a measure on the state of ReLU-activated neurons, to be used as an analysis tool to get a better understanding for the obtained sparse ANN models.\\
The rest of this paper is organized as follows. Sec.~\ref{sec:sota} reviews the importance of network pruning and the most relevant literature. Next, in Sec.~\ref{sec:minima} we discuss the relevant literature around local properties of minima for ANN models. Then, in Sec.~\ref{sec:entropy} we propose PSP-entropy, a metric to measure how much a neuron specializes in identifying features belonging to a sub-set of classes learned at training time. Sec.~\ref{sec:results1} provides our findings on the properties for sparse architectures and finally, in Sec.~\ref{sec:conclusion}, we draw the conclusions and identify further directions for future research.
\section{State of the art pruning techniques}
\label{sec:sota}
In the literature it is possible to find a large number of pruning approaches, some old-fashioned~\cite{lecun1990optimal} and others more recent~\cite{han2015learning,li2016pruning,molchanov2017variational}. Among the latter, many sub-categories can be identified. Ullrich~\emph{et~al.} introduce what they call \emph{soft weight sharing}, through which is possible to introduce redundancy in the network and reduce the amount of stored parameters~\cite{ullrich2017soft}. Other approaches are based on parameters regularization and pruning: for example, Louizos~\emph{et~al.} use an $L_0$ proxy regularization; Tartaglione~\emph{et~al.}, instead, define the importance of a parameter via a sensitivity measure used as regularization~\cite{tartaglione2018sensitivity}. Other approaches are dropout-based, like \emph{sparse variational dropout}, proposed by Molchanov~\emph{et~al.}, leveraging on a bayesian interpretation of Gaussian dropout and promoting sparsity in the ANN model~\cite{molchanov2017variational}.\\
Overall, most of the proposed pruning techniques can be divided in two macro classes. The first is defined by approaches based on \emph{gradual pruning}~\cite{louizos2017learning,lobster,zhu2017prune}, in which the network is, at the same time, trained and pruned following some heuristic approach, spanning a large number of pruning iterations. One among these, showing the best performances, is LOBSTER, where parameters are gradually pruned according to their local contribution to the loss~\cite{lobster}. The second class, instead, includes all the techniques based on \emph{one-shot pruning}~\cite{frankle2018lottery,han2015learning,luo2017thinet}: here the pruning procedure consists of three stages:
\begin{enumerate}
    \item a large, over-parametrized network is normally trained to completion;
    \item the network is then pruned using some kind of heuristic (e.g. magnitude thresholding) to satisfaction (the percentage of remaining parameters is typically an hyper-parameter);
    \item the pruned model is further fine-tuned to recover the accuracy lost due to the pruning stage.
\end{enumerate}
A recent work by Frankle~and~Carbin~\cite{frankle2018lottery} termed the \emph{lottery ticket hypothesis}, which is having a large impact on the research community.
They claim that from an ANN, early in the training, it is possible to extract a sparse sub-network on a one-shot fashion: such sparse network, when trained, can match the accuracy of the original model. In a follow-up, Renda~\emph{et~al.} propose a retraining approach that replaces the fine-tuning step with \emph{weight rewinding}: after pruning, the remaining parameters are reset to their initial values and the pruned network is trained again. They also argue that using the initial weights values is fundamental to achieve competitive performance, which is degraded when starting from a random initialization~\cite{renda2020comparing}.\\
On the other hand, Liu~\emph{et~al.} show that, even when retraining a pruned sub-network using a new random initialization, they are able to reach an accuracy level comparable to its dense counterpart; challenging one of the conjectures proposed alongside the lottery ticket hypothesis~\cite{liu2018rethinking}.\\
In our work we try to shed some light on this discussion, comparing state-of-the-art one-shot pruning to gradual pruning.
\section{Local properties of minima}
\label{sec:minima}
In the previous section we have explored some of the most relevant pruning strategies. All of them rely on state-of-the-art optimization strategies: applying very simple optimizing heuristics to minimize the loss function, like for example SGD~\cite{bottou2010large,zinkevich2010parallelized}, it is nowadays possible to succeed in training ANNs on huge datasets. Theoretically speaking, this is the ``miracle'' of deep learning, as the dimensionality of the problem is huge (indeed, these problems are typically over-parametrized, and the dimensionality can be efficiently reduced~\cite{tartaglione2018sensitivity}). Furthermore, minimizing non-convex objective functions is typically supposed to make the trained architecture stuck into local minima. However, the empirical evidence shows that something else is happening under the hood: understanding it is in general critical.\\
Goodfellow~\emph{et~al.} pioneered the problem of understanding why deep learning works. In particular, they observed there is essentially no loss barrier between a generic random initialization for the ANN model and the final configuration~\cite{goodfellow2014qualitatively}. This phenomena has also been observed on larger architectures by Draxler~\emph{et~al.}~\cite{draxler2018essentially}. These works lay as basis for the ``lottery ticket hypothesis'' papers. However, a secondary yet relevant observation in~\cite{goodfellow2014qualitatively} stated that there is a loss barrier between different ANN configurations showing similar generalization capabilities. Later, it was shown that typically a low loss path between well-generalizing solutions to the same learning problem can be found~\cite{tartaglione2019take}. From this brief discussion it is evident that a general approach on how to better characterize such minima has yet to be found.\\
Keskar~\emph{et~al.} showed why we should prefer small batch methods to large batch ones: they correlate the stochasticity introduced by small-batch methods to the sharpness of the reached minimum~\cite{keskar2016large}. In general, they observe that the larger the batch, the sharper the reached minimum. Even more interestingly, they observe that the sharper the minimum, the worse the generalization of the ANN model. In general, there are many works supporting the hypothesis that flat minima generalize well, and this has been also the strength for a significant part of the current research~\cite{chaudhari2016entropy,keskar2016large}. However, in general this does not necessarily mean that no sharp minimum generalizes well, as we will see in Sec.~\ref{sec:sharp}.
\section{Towards a deeper understanding: an entropy-based approach}
\label{sec:entropy}
In this section we propose PSP-entropy, a metric to evaluate the dependence of the output for a given neuron in the ANN model to the target classification task. The proposed measure will allow us to better understand the effect of pruning.

\subsection{Post-synaptic potential}
Let us define the output of the given $i$-th neuron at the $l$-th layer as
\begin{equation}
    y_{l,i} = \varphi\left[ f(\mathbf{y}_{l-1}, \mathbf{\theta}_{l,i}\right])
\end{equation}
where $\mathbf{y}_{l-1}$ is the input of such neuron, $\mathbf{\theta}_{l,i}$ are the parameters associated to it, $f(\cdot)$ is some affine function and $\varphi(\cdot)$ is the activation function, we can define its \emph{post-synaptic potential} (PSP)~\cite{tartaglione2019post} as
\begin{equation}
    z_{l,i} = f(\mathbf{y}_{l-1}, \mathbf{\theta}_{l,i})
\end{equation}
Typically, deep models are ReLU-activated: here on, let us consider the activation function for all the neurons in hidden layers as $\varphi(\cdot)=\relu(\cdot)$. Under such assumption it is straightforward to identify two distinct regions for the neuron activation:
\begin{itemize}
    \item $z_{l,i}\leq 0$: the output of the neuron will be exactly zero $\forall z_{l,i} \leq 0$;
    \item $z_{l,i} > 0$: there is a linear dependence of the output to $z_{l,i}$.
\end{itemize}
Hence, let us define
\begin{equation}
    \varphi'(z) = \left\{
    \begin{array}{ll}
        0 &\ \ \ z \leq 0\\
        1 &\ \ \ z > 0
    \end{array}
    \right .
\end{equation}
Intuitively, we understand that if two neurons belonging to the same layer, for the same input, share the same $\varphi'(z)$, then they are linearly-mappable to one equivalent neuron:
\begin{itemize}
    \item $z_{l,i}\leq 0$, $z_{l,j}\leq 0$: one of them can be simply removed;
    \item $z_{l,i} > 0$, $z_{l,j} > 0$: they are equivalent to a linear combination of them.
\end{itemize}
In this work we are not interested in using this approach towards structured pruning: there are many works in the literature which tackle this issue using efficient proxies. In the next section we are going to formulate a metric to evaluate the degree of disorder in the post synaptic potentials. The aim of such measure will be to have an analytical tool to give us a broader understanding on the behavior of the neurons in sparse architectures.

\subsection{PSP-entropy for ReLU-activated neurons}
In the previous section we have recalled the concept of post-synaptic potential. Some interesting concepts have been also introduced for ReLU-activated networks: we can use its value to approach the problem of \emph{binning} the state of a neuron, according to $\varphi'(z_{l,i})$. Hence, we can construct a binary random process that we can rank according to its entropy. To this end, let us assume we set as input of our ANN model two different patterns, $\mu_{c,1}$ and $\mu_{c,2}$, belonging to the same class $c$ (for those inputs, we aim at having the same target at the output of the ANN model). Let us consider the PSP $z_{l,i}$ (where $l$ is an hidden layer): 
\begin{itemize}
    \item if $\varphi'(z_{l,i}|\mu_{c,1})$ = $\varphi'(z_{l,i}|\mu_{c,2})$ we can say there is \emph{low PSP entropy};
    \item if $\varphi'(z_{l,i}|\mu_{c,1})$ $\neq$ $\varphi'(z_{l,i}|\mu_{c,2})$ we can say there is \emph{high PSP entropy}.
\end{itemize}
We can model an entropy measure for PSP:
\begin{equation}
    \label{eq:Hsimple}
    H(z_{l,i}| c) = -\sum_{t=\{0,1\}} p\left[\varphi'(z_{l,i})=t| c\right]\cdot \log_2\left\{p\left[\varphi'(z_{l,i})=t| c\right]\right\}
\end{equation}
where $p[\varphi'(z_{l,i})=t| c]$ is the probability $\varphi'(z_{l,i}) = t$ when presented an input belonging to the $c$-th class. Since we typically aim at solving a multi-class problem, we can model an overall entropy for the neuron as
\begin{equation}
    \label{eq:H1}
    H(z_{l,i}) = \sum_{c} H(z_{l,i}| c)
\end{equation}
It is very important to separate the contributions of the entropy according to the $c$-th target class since we expect the neurons to catch relevant features being highly-correlated to the target classes. Eq.~\eqref{eq:H1} provides us very important information towards this end: the lower its value the more it specializes for some specific classes.\\
The formulation in~\eqref{eq:H1} is very general and it can be easily extended to higher-order entropy, i.e. entropy of sets of neurons whose state correlates for the same classes. Now we are ready to use this metrics to shed further light to the findings in Sec.~\ref{sec:results1}.

\section{Experiments}
\label{sec:results1}
For our test, we have decided to compare the state-of-the-art one-shot pruning proposed by Frankle~and~Carbin~\cite{frankle2018lottery} to one of the top-performing gradual pruning strategies, LOBSTER~\cite{lobster}. Towards this end, we first obtain a sparse network model using LOBSTER;
the non-pruned parameters 
are then re-initialized to their original values, according to the lottery ticket hypothesis~\cite{frankle2018lottery}. Our purpose here is to determine whether the lottery ticket hypothesis applies also to the sparse models obtained using high-performing gradual pruning strategies.\\
As a second experiment, we want to test the effects of different, random initialization while keeping the achieved sparse architecture. According to Liu~\emph{et al.}, this should lead to similar results to those obtained with the original initialization~\cite{liu2018rethinking}. Towards this end, we tried $10$ different new starting configurations. As a last experiment, we want to assess how important is the structure originating from the pruning algorithm in reaching competitive performances after re-initialization: for this purpose, we randomly define a new pruned architecture with the same number of pruned parameters as those found via LOBSTER. Also in this case, $10$ different structures have been tested.\\
We decided to experiment with different architectures and datasets commonly employed in the relevant literature: LeNet-300 and LeNet-5-caffe trained on MNIST, LeNet-5-caffe trained on Fashion-MNIST~\cite{xiao2017fashionmnist} and ResNet-32 trained on CIFAR-10.\footnote{\url{https://github.com/akamaster/pytorch_resnet_cifar10}} For all our trainings we used the SGD optimization method with standard hyper-parameters and data augmentation, as defined in the papers of the different compared techniques~\cite{frankle2018lottery,liu2018rethinking,lobster}.

\subsection{One-shot vs gradual pruning}
\label{sec:1svsit}
\begin{figure*}
    \centering

    \begin{subfigure}{0.75\textwidth}
        \centering
        \includegraphics[width=1\linewidth]{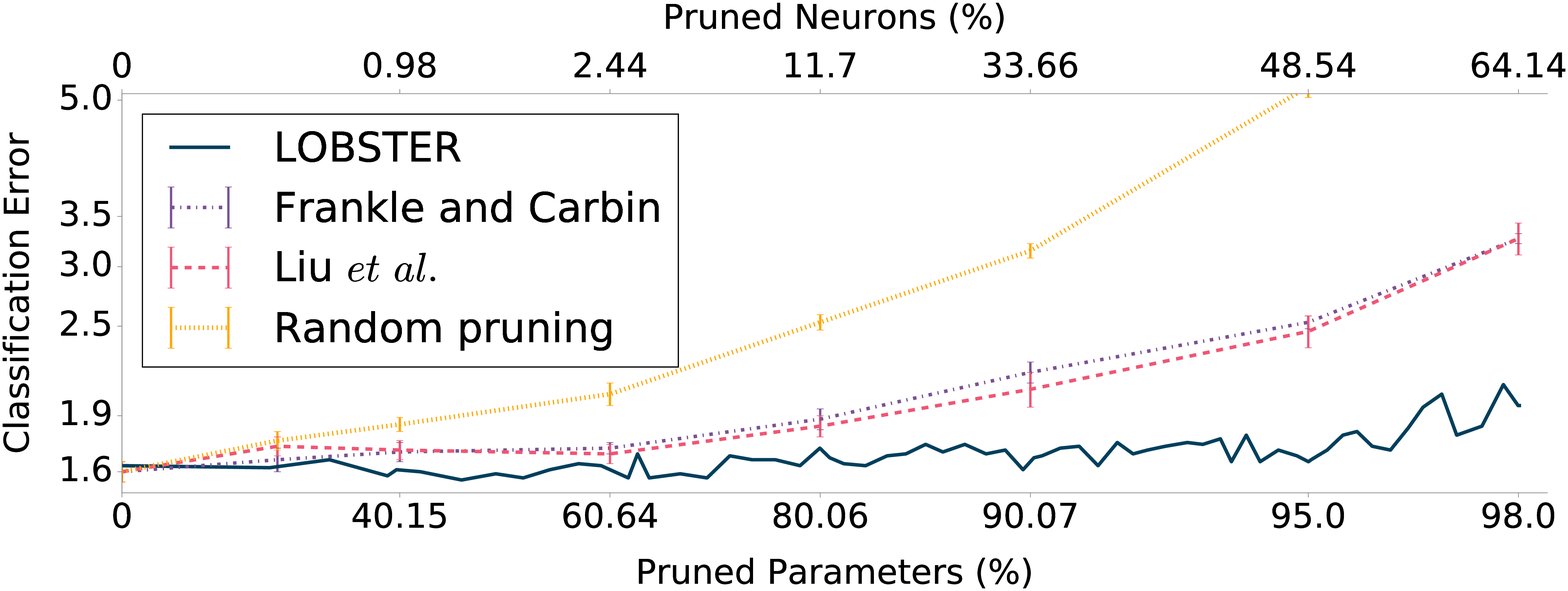}
        \caption{~}
        \label{fig::ln300_LOB}
    \end{subfigure}
    
    \vspace{6 mm}

    \begin{subfigure}{0.75\textwidth}
        \centering
        \includegraphics[width=1\linewidth]{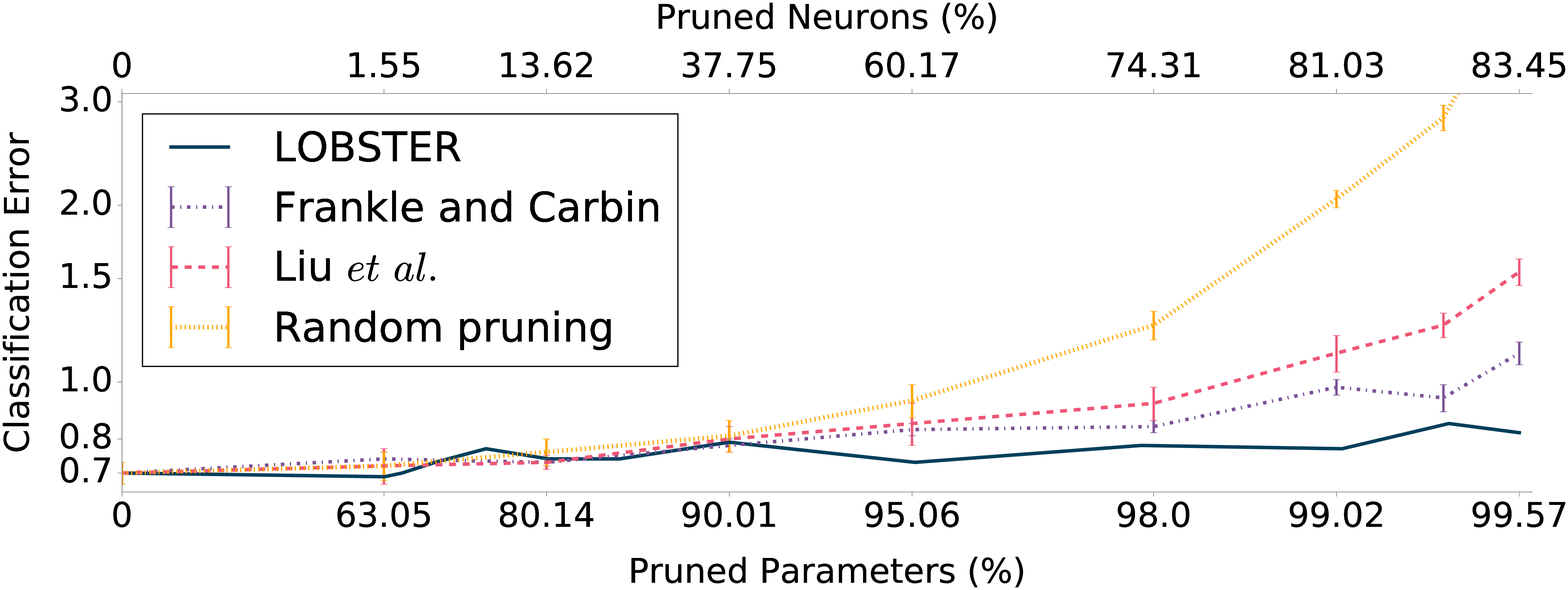}
        \caption{~}
        \label{fig::ln5_M_LOB}
    \end{subfigure}
    
    \vspace{6 mm}

    \begin{subfigure}{0.75\textwidth}
        \centering
        \includegraphics[width=1\linewidth]{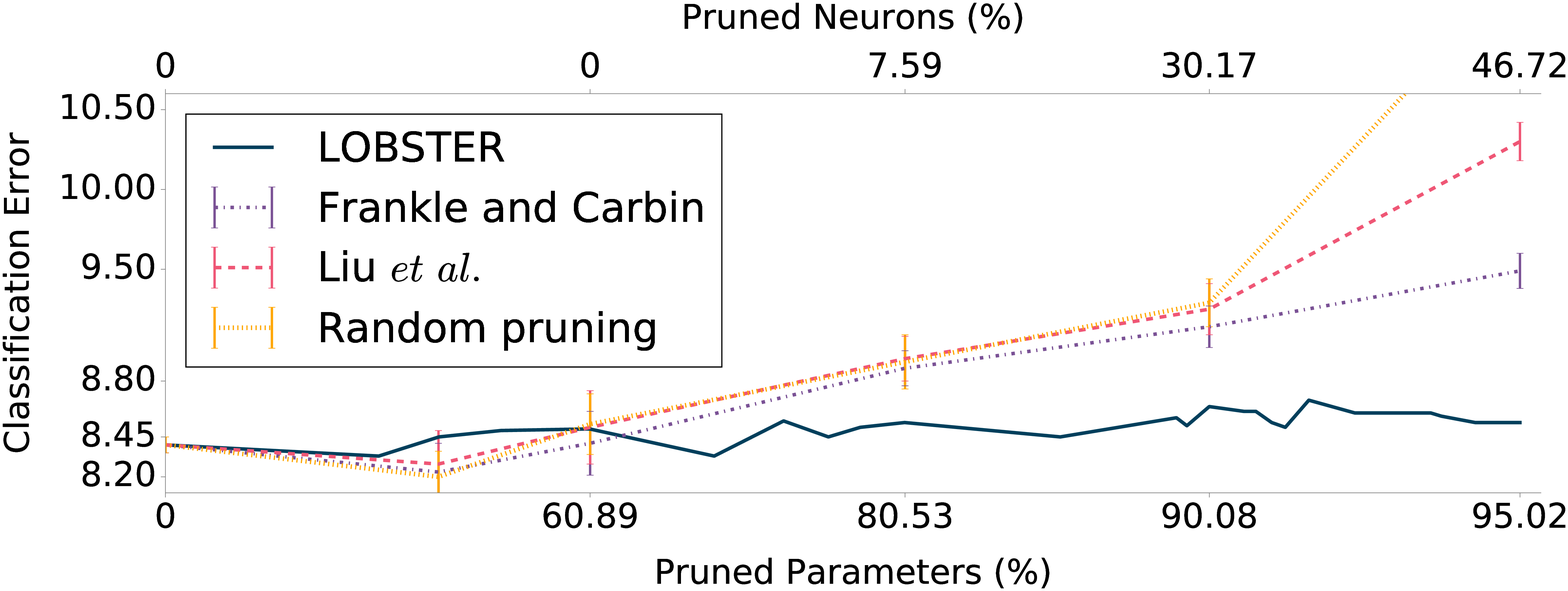}
        \caption{~}
        \label{fig::ln5_F_LOB}
    \end{subfigure}
    
    \vspace{6 mm}

    \begin{subfigure}{0.75\textwidth}
        \centering
        \includegraphics[width=1\linewidth]{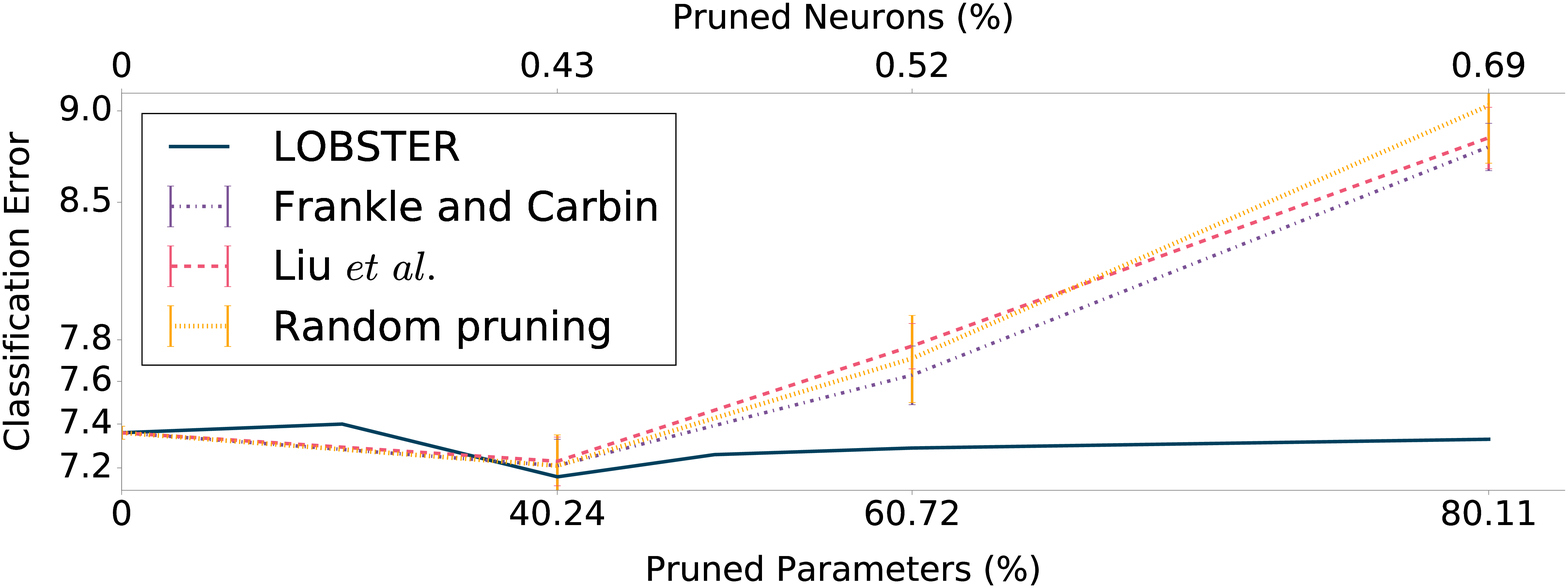}
        \caption{~}
        \label{fig::res32_LOB}
    \end{subfigure}
    
    \caption{Test set error for different compression rates: LeNet-300~(a) trained on MNIST, LeNet-5 trained on MNIST~(b), LeNet-5 trained on Fashion-MNIST~(c) and ResNet-32 trained on CIFAR-10~(d).}
    \label{fig::lobstervslottery_1}
\end{figure*}
In Fig.~\ref{fig::lobstervslottery_1} we show, for different percentages of pruned parameters, a comparison between the test accuracy of models pruned using the LOBSTER technique and the models retrained following the approaches we previously defined.\\
We can clearly identify a low compression rate regime in which the re-initialized model is able to recover the original accuracy, validating the lottery ticket hypothesis. On the other hand, when the compression rate rises (for example when we remove more than 95\% of the LeNet-300 model's parameters, as observed in Fig.~\ref{fig::ln300_LOB}), the re-training approach strives in achieving low classification errors.\\
As one might expect, other combinations of dataset and models might react differently. For example, LeNet-300 is no longer able to reproduce the original performance when composed of less then 5\% of the original parameters. On the other hand, LeNet-5, when applied on MNIST, is able to achieve an accuracy of around 99.20\% even when 98\% of its parameters are pruned away (Fig.~\ref{fig::ln5_M_LOB}). This does not happen when applied on a more complex dataset like Fashion-MNIST, where removing 80\% of the parameters already leads to performance degradation (Fig.~\ref{fig::ln5_F_LOB}). Such a gap becomes extremely evident when we re-initialize an even more complex architecture like ResNet-32 trained on CIFAR-10 (Fig.~\ref{fig::res32_LOB}).\\
From the reported results, we observe that the original initialization is not always important: the error gap between a randomly initialized model and a model using the original weights' values is minor, with the latter being slightly better. Furthermore, they both fail in recovering the performance for high compression rates.

\subsection{Sharp minima can also generalize well}
\label{sec:sharp}
\begin{figure*}
    \begin{center}
    \begin{subfigure}{0.48\textwidth}
        \centering
        \includegraphics[width=\textwidth]{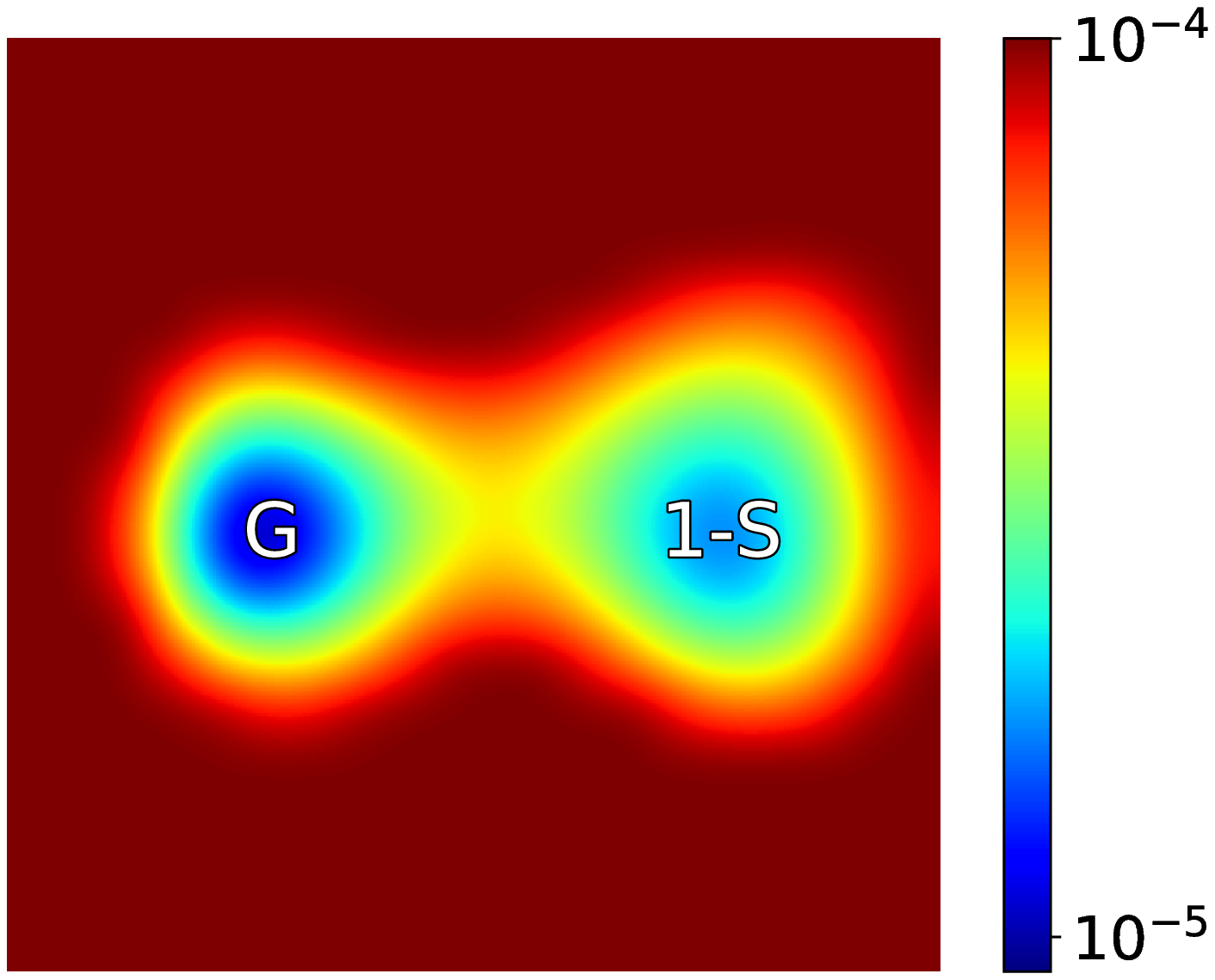}
        \caption{~}
        \label{fig::ln5_loss_train}
    \end{subfigure}
    \begin{subfigure}{0.48\textwidth}
        \centering
        \includegraphics[width=\textwidth]{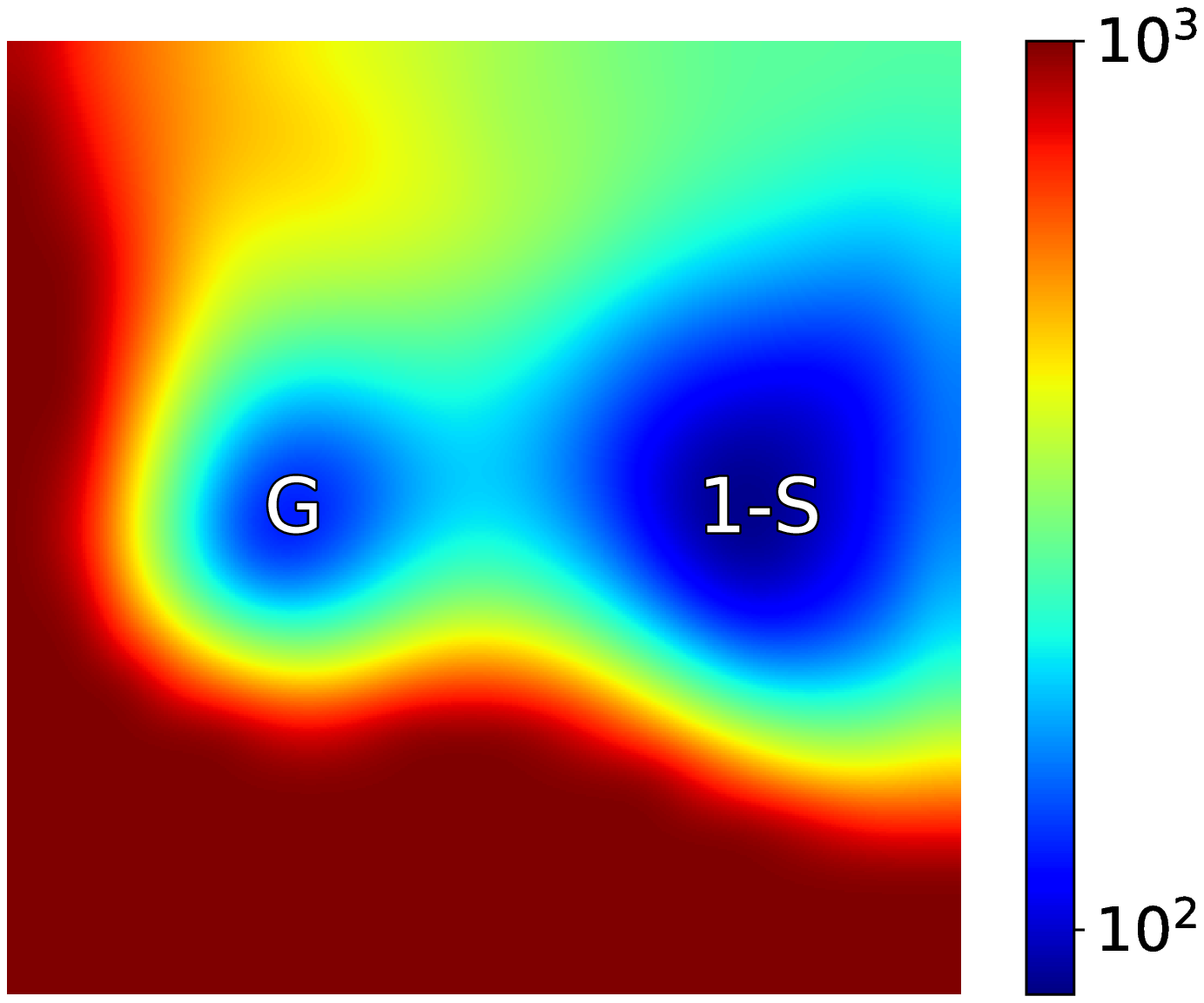}
        \caption{~}
        \label{fig::ln5_hess_train}
    \end{subfigure}
    \end{center}
    
    \caption{Results of LeNet-5 trained on MNIST with the highest compression (99.57\%): (a) plots loss in the training set and (b) plots the top-5 largest hessian eigenvalues. G is the solution found with gradual learning while 1-S is the best one-shot solution (Frankle and Carbin).}
    \label{fig::heatmap-ln5}
\end{figure*}

\noindent In order to study the sharpness of local minima, let us focus, for example, on the results obtained on LeNet-5 trained on MNIST. We choose to focus our attention on this particular ANN model since, according to the state-of-the-art and coherently to our findings, we observe the lowest performance gap between gradual and one-shot pruning (as depicted in Fig.~\ref{fig::ln5_M_LOB}); hence, it is a more challenging scenario to observe qualitative differences between the two approaches. However, we remark that all the observations for such a case apply also to the other architectures/datasets explored in Sec.~\ref{sec:1svsit}.\\
In order to obtain the maps in Fig.~\ref{fig::heatmap-ln5}, we follow the approach proposed by~\cite{goodfellow2014qualitatively} and we plot the loss for the ANN configurations between two reference ones: in our, case, we compare a solution found with gradual pruning (G) and one-shot (1-S). Then, we take a random orthogonal direction to generate a 2D map. Fig.~\ref{fig::ln5_loss_train} shows the loss on the training set between iterative and one-shot pruning for the highest compression rate (99.57\% of pruned parameters as shown in Fig.~\ref{fig::ln5_M_LOB}). According to our previous findings, we see that iterative pruning lies in a lower loss region. Here, we show also the plot of the top-5 Hessian eigenvalues (all positive), in Fig.~\ref{fig::ln5_hess_train}, using the efficient approach proposed in~\cite{hessian-eigenthings}. Very interestingly, we observe that the solution proposed by iterative pruning lies in a narrower minimum than the one found using the one-shot strategy, despite generalizing slightly better. With this, we do not claim that narrower minima generalize well: gradual pruning strategies enable access to a \emph{subset of well-generalizing narrow minima}, showing that not all the narrow minima generalize worse than the wide ones. This finding raises warnings against second order optimization, which might favor the research of flatter, wider minima, ignoring well-generalizing narrow minima. These non-trivial solutions are naturally found using gradual pruning which cannot be found using one-shot approaches, which on the contrary focus their effort on larger minima. In general, the sharpness of these minima explains why, for high compression rates, re-training strategies fail in recovering the performance, considering that it is in general harder to access this class of minima.

\subsection{Study on the post synaptic potential}
\begin{figure}
    \centering
    \includegraphics[width=0.4\textwidth]{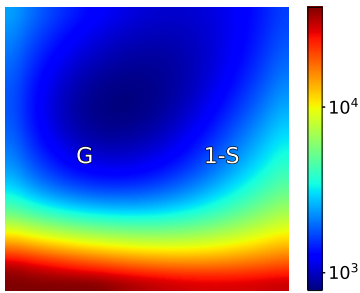}
    \caption{L2 norm of PSP values for LeNet-5 trained on MNIST with 99.57\% of pruned parameters.}
    \label{fig::z}
\end{figure}
\noindent In Sec.~\ref{sec:sharp} we have observed that, as a result, iterative strategies focus on well-generalizing sharp minima. Is there something else yet to say about those?\\
Let us inspect the average magnitude values of the PSPs for the different found solutions: towards this end, we could plot the average of their L2 norm values ($z^2$). As a first finding, gradually-pruned architectures naturally have lower PSP L2-norm values, as we observe in Fig.~\ref{fig::z}. None of the used pruning strategies explicitly minimize the term in $z^2$: they naturally drive the learning towards such regions. However, the solution showing better generalization capabilities shows lower $z^2$ values. Of course, there are regions with even lower $z^2$ values; however, according to Fig.~\ref{fig::ln5_loss_train}, they should be excluded since they correspond to high-loss values (not all the low $z^2$ regions are low-loss).
\begin{figure*}
    \begin{center}
    \begin{subfigure}{0.48\textwidth}
        \centering
        \includegraphics[width=\textwidth]{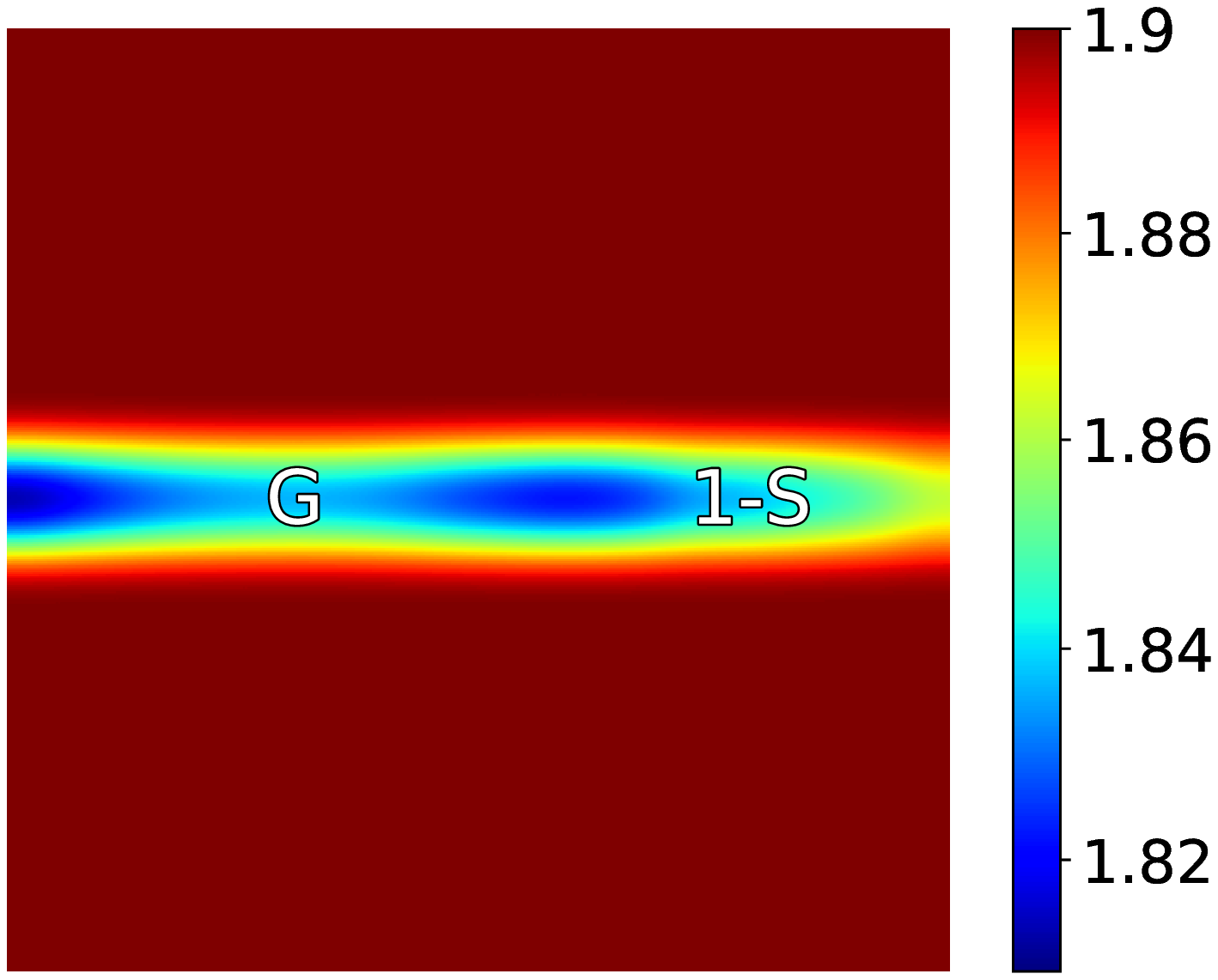}
        \caption{~}
        \label{fig::H1}
    \end{subfigure}
      ~
    \begin{subfigure}{0.48\textwidth}
        \centering
        \includegraphics[width=\textwidth]{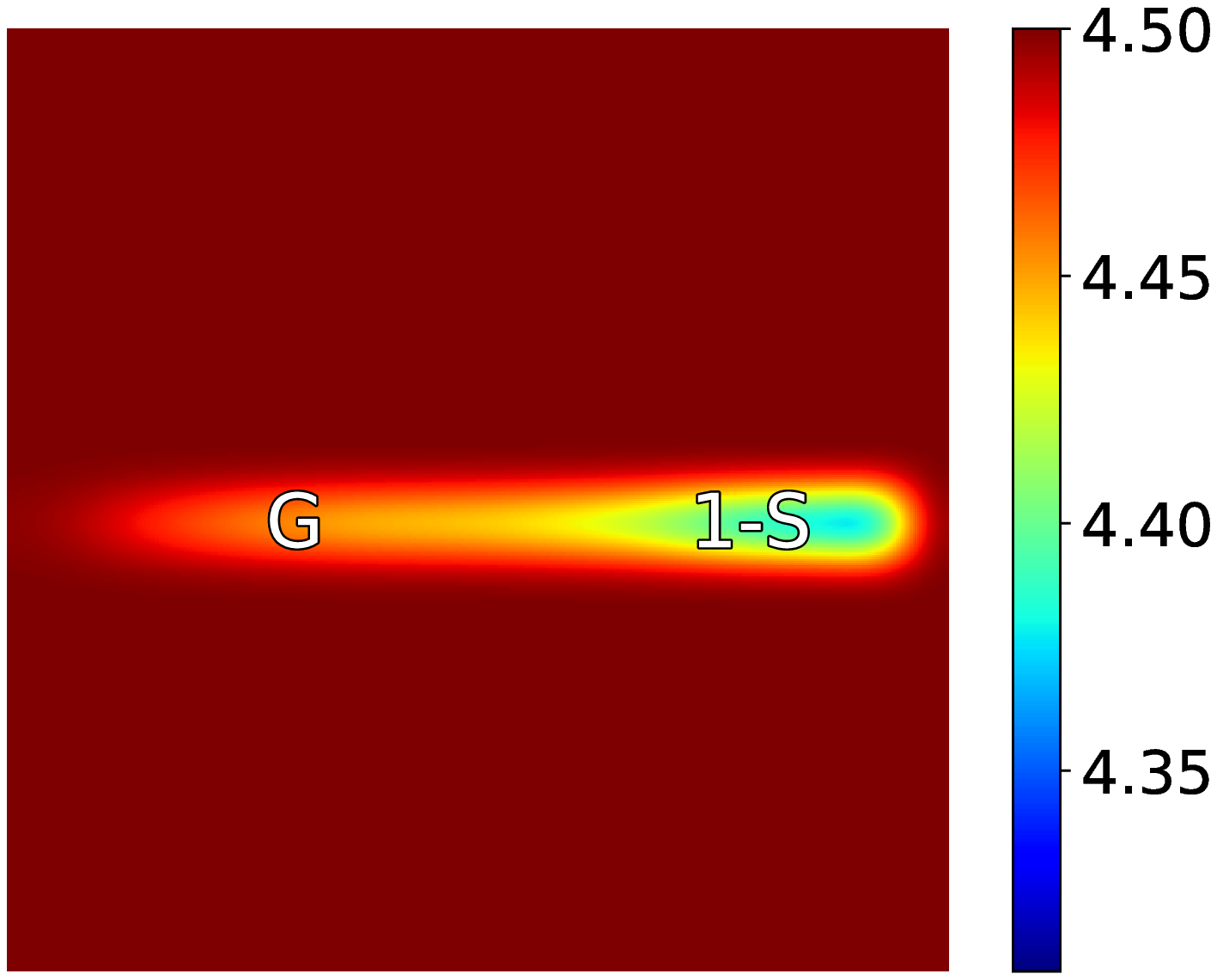}
        \caption{~}
        \label{fig::H2}
    \end{subfigure}
    \end{center}
    \caption{Results on LeNet-5 trained on MNIST with 99.57\% of pruned parameters. (a) plots the first order PSP-entropy, while (b) shows the second-order PSP entropy.}
    \label{fig::entropy}
\end{figure*}
If we look at the PSP-entropy formulated in~\eqref{eq:H1}, we observe something interesting: gradual and one-shot pruning show comparable first-order entropies, as shown in Fig.~\ref{fig::H1}.\footnote{the source code for PSP-entropy is available at \url{https://github.com/EIDOSlab/PSP-entropy.git}} It is interesting to see that there are also lower entropy regions which however correspond to higher loss values, according to Fig.~\ref{fig::ln5_loss_train}. When we move to higher-order entropies, something even more interesting happens: gradual pruning shows higher entropy than one-shot, as depicted in Fig.~\ref{fig::H2} (displaying the second order entropy). In such a case, having a lower entropy means having more groups of neurons specializing to specific patterns which correlate to the target class; on the contrary, having higher entropy yet showing better generalization performance results in having more general features, more agnostic towards a specific class, which still allow a correct classification performed by the output layer. This counter-intuitive finding has potentially-huge applications in transfer learning and domain adaptation, where it is critical to extract more general features, not very specific to the originally-trained problem.

\section{Conclusion}
\label{sec:conclusion}
In this work we have compared one-shot and gradual pruning on different state-of-the-art architectures and datasets. In particular, we have focused our attention in understanding potential differences and limits of both approaches towards achieving sparsity in ANN models.\\
We have observed that one-shot strategies are very efficient to achieve moderate sparsity at a lower computational cost. However, there is a limit to the maximum achievable sparsity, which can be overcome using gradual pruning. The highly-sparse architecture, interestingly, focus on a subset of sharp minima which are able to generalize well, which pose some questions to the potential sub-optimality of second-order optimization in such scenarios. This explains why we observe that one-shot strategies fail in recovering the performance for high compression rates. More importantly, we have observed, contrarily to what it could be expected, that highly-sparse gradually-pruned architectures are able to extract general features non-strictly correlated to the trained classes, making them unexpectedly, potentially, a good match for transfer-learning scenarios.\\
Future works include a quantitative study on transfer-learning for sparse architectures and PSP-entropy maximization-based learning.

%
%
\bibliographystyle{splncs04}
\bibliography{mybibliography}
\end{document}